\documentclass{article}
\usepackage{spconf,amsmath,graphicx}

\title{Learning Optical Flow via Dilated Networks and Occlusion Reasoning}
\name{Yi Zhu and Shawn Newsam}

\address{University of California, Merced \\
	5200 N Lake Rd, Merced, CA, US  \\
	\{yzhu25, snewsam\}@ucmerced.edu }

\begin{document}
%
\maketitle
\begin{abstract}
Despite the significant progress that has been made on estimating optical flow recently, most estimation methods, including classical and deep learning approaches, still have difficulty with multi-scale estimation, real-time computation, and/or occlusion reasoning. In this paper, we introduce dilated convolution and occlusion reasoning into unsupervised optical flow estimation to address these issues. The dilated convolution allows our network to avoid upsampling via deconvolution and the resulting gridding artifacts. Dilated convolution also results in a smaller memory footprint which speeds up interference. The occlusion reasoning prevents our network from learning incorrect deformations due to occluded image regions during training. Our proposed method outperforms state-of-the-art unsupervised approaches on the KITTI benchmark. We also demonstrate its generalization capability by applying it to action recognition in video.
\end{abstract}

\section{Introduction}
\label{sec:intro}
Optical flow is valuable for image sequence analysis due to its ability to encode motion. Significant progress has been made on the estimation of optical flow over the past few years. Classical approaches are typically based on variational models and solved as an energy minimization process \cite{Horn_Schunck,opticalFlowWarp2004,brox_flow_matching_11}. They remain the top performers on a number of evaluation benchmarks. Most, however, are too slow to be used in real time applications. By contrast, convolutional neural network (CNN) based methods formulate optical flow estimation as a learning task and can reduce inference time to fractions of a second. However, despite their increased accuracy, most flow estimation methods, including classical and CNN approaches, still having difficulty with multi-scale estimation, gridding artifacts, real-time computation, and/or occlusion reasoning.

We focus on CNN based approaches in this work due to their efficiency. FlowNet \cite{flownet} was the first work to directly learn optical flow given an image pair using CNNs. To deal with multi-scale, FlowNet2 \cite{flownet2} proposed separate streams to encode large and small scale displacement which are then fused using a refinement network. Although FlowNet2 achieves good performance, the memory footprint of the model is high due to the five separate networks similar to FlowNet. Another approach, SPyNet \cite{spynet_16}, instead adopts spatial pyramids to output optical flow at multiple resolutions. Its network is $96\%$ smaller than that of FlowNet and results in increased performance. These are all supervised methods, however, that require ground truth optical flow during training which is only available for synthetic data. The transferability from synthetic to real domains remains an open question.

Unsupervised \cite{AhmadiICIP2016,jasonUnsup2016,hidden_zhu_17} or semi-supervised \cite{guided_zhu_2017,Lai-NIPS-2017} approaches which do not require ground truth flow have thus been developed. These methods usually guide the learning using an image reconstruction loss based on a brightness constancy assumption. Although this allows for unlimited training data, the performance is limited by such a loss function. Indeed, these unsupervised approaches tend to lag far behind their supervised counterparts on standard benchmarks. One reason for this is that the loss is based on photo-consistency error which is only meaningful when there is no occlusion. Without explicit occlusion reasoning, the back-propagated gradients are incorrect and degrade the training. 

Two concurrent works \cite{occlusionFlow_wang17,unflow_aaai2018} consider occlusion explicitly for estimating optical flow using CNNs. The intuition is based on a forward-backward consistency assumption. That is, for non-occluded pixels, the forward flow should be the inverse of the backward flow between image pairs. During training, gradients are calculated from non-occluded regions only. 
Our work is most similar to \cite{unflow_aaai2018} in terms of unsupervised learning and occlusion reasoning, but differs in several ways: (1) We adopt dilated operations for the last few convolutional groups, which leads to high resolution feature maps throughout the network and improved multi-scale handling. (2) We incorporate dense connections in the network instead of sparse skip connections. This strategy helps to capture thin structures and small object displacements. And, (3) our network completely avoid upsampling via deconvolution and thus greatly reduces gridding artifacts. Our proposed framework outperforms state-of-the-art unsupervised approaches on standard benchmarks, and is shown to generalize well to action recognition.

\begin{figure*}[t]
	\centering
	\includegraphics[width=1.0\linewidth,trim=0 0 0 0,clip]{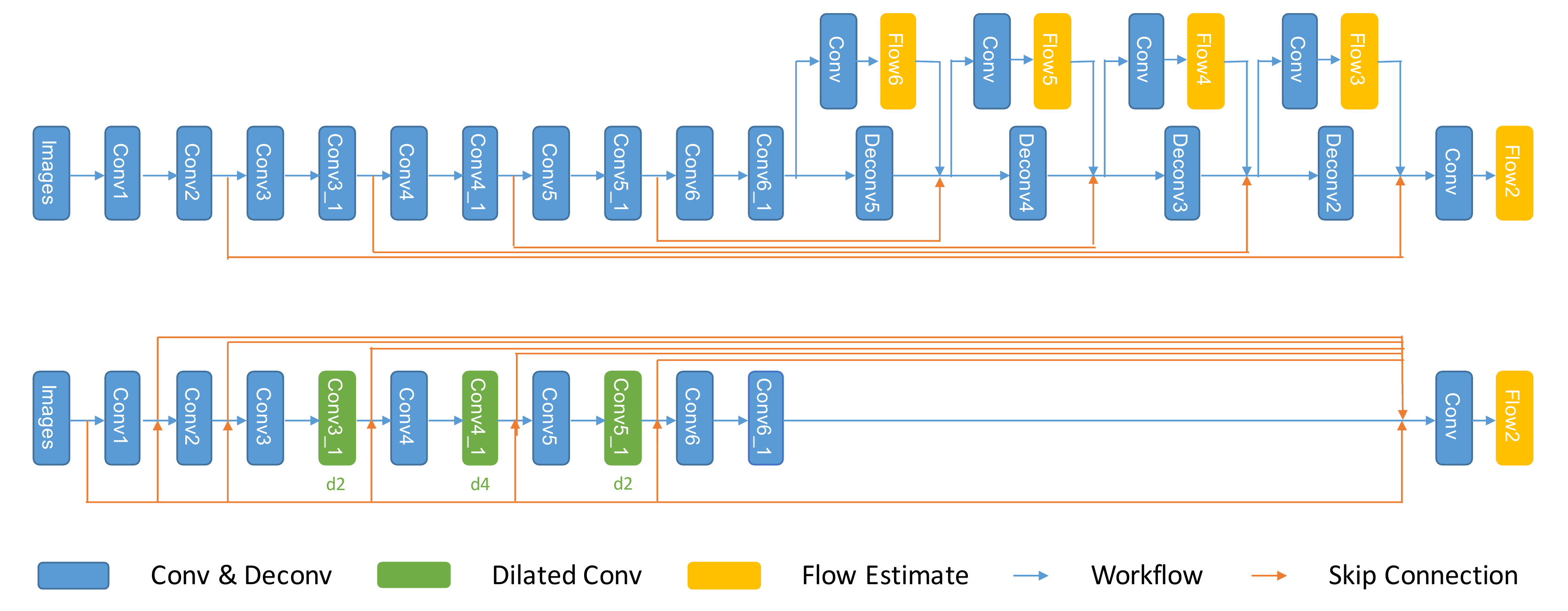}
	\vspace{-4ex}
	\caption{Upper: original FlowNetS. Bottom: our unsupervised learning framework based on dilated convolution. For the three dilated convolutions (green), d2 and d4 denote a dilation factor of 2 and 4, respectively. The figure is best viewed in color. }
	\label{fig:overview}
\end{figure*}

\section{Method}
\label{sec:method}

Given consecutive frames, $I_{1}$ and $I_{2}$, our goal is to learn a model that can predict the per-pixel motion field $(u, v)$ between the two images. $u$ and $v$ are the displacements in the horizontal and vertical direction.

\subsection{Dilated Networks}
\label{sec:dilate}

Conventional CNNs apply progressive downsampling to extract high-level semantic representations and to reduce computational costs. While this has proven effective, the reduced resolution results in a loss of spatial information that can limit detailed image understanding tasks such as semantic segmentation, optical flow or depth estimation, etc. It is preferable to preserve the spatial resolution throughout the network. 

One popular way to increase the resolution of the network output is to follow the convolutional stage with several deconvolutional layers \cite{flownet} (upper network in Fig. \ref{fig:overview}). However, this has a number of drawbacks including more parameters to learn, gridding artifacts, and additional parameter selection for the multi-scale losses.

In contrast to standard convolution and pooling layers, dilated convolution \cite{dilate_Yu_iclr16} increases the receptive view (scale) of the network exponentially without resulting in a loss of resolution or coverage. Filters are dilated so that they operate at different scales. (Details about dilated convolution can be found in the supplemental materials.) Dilated convolution allows: (1) the detection of finer details by processing the inputs at higher resolutions; (2) a broader view of the input so as to capture more contextual information; (3) faster inference with fewer parameters. These benefits align with our task at hand and so we incorporate dilated convolution into our flow estimation framework. 

The bottom part of Fig. \ref{fig:overview} shows our proposed network. Our encoder structure is similar to that of FlowNetS but incorporates dilated convolution in the later convolutional groups (i.e., conv3$\_$1, conv4$\_$1, conv5$\_$1). This preserves the resolution of the feature maps from layer conv3 through to the end of the network and allows our output, Flow2, to be the same size as that of FlowNetS without the need for the deconvolutional layers. This has several major benefits: our network has around half the number of parameters as FlowNetS, does not suffer from gridding artifacts, and does not require time-consuming parameter tuning for the multi-scale loses. Indeed, our network converges three times faster during training and runs twice as fast during inference as FlowNetS.

In addition, inspired by \cite{densenet_16,densenet_zhu_icip2017}, the feature maps in our network are densely connected. This preserves the high frequency image details. When concatenating feature maps of different sizes, we simply downsample the larger one or upsample the smaller one. We also route the original RGB image pairs to each convolutional group for guided filtering \cite{guided_filter_He_eccv10}. 

\subsection{Degridding}
\label{sec:degrid}

In \cite{flownet,flownet2,jasonUnsup2016}, the use of deconvolution can often cause gridding artifacts, commonly known as ``checkerboard artifacts'' \cite{odena2016deconvolution}. In particular, the deconvolutions have uneven overlap when the kernel size is not divisible by the stride which can result in gridding. While this can be avoided by constraining the kernel size to be divisible by the stride, artifacts at a variety of scales can still occur due to the compounding effect of the stacked deconvolutions. We thus employ dilated convolution to preserve the resolution while avoiding the upsampling of deconvolution and its artifacts.

However, dilated convolution can produce its own gridding artifacts \cite{drn_Yu_cvpr17}. This occurs when a feature map has  higher frequency content than the sampling rate of the dilated convolution. To prevent this, we add additional dilated convolutional layers to the end of the network with progressively smaller dilation factors. As shown in Fig. \ref{fig:overview}, after the 4-dilated layer conv4$\_$1, we add a 2-dilated layer conv5$\_$1 followed by a 1-dilated layer conv6$\_$1. This has an effect similar to that of using appropriately tuned filters to remove aliasing artifacts, and results in smoother flow estimates.

\subsection{Unsupervised Motion Estimation}
\label{sec:unsup}

Most unsupervised methods \cite{jasonUnsup2016} treat optical flow estimation as an image reconstruction problem. The intuition is that if we can use the predicted flow and the next frame to reconstruct the previous frame, our model has learned a useful representation of the underlying motion. Let us denote the \emph{reconstructed} previous frame as $I_{1}^{\prime}$. The goal then is to minimize the photometric error between the true previous frame $I_{1}$ and the reconstructed previous frame $I_{1}^{\prime}$: 
\begin{equation}
L_{\text{reconst}} = \frac{1}{N} \sum_{i, j}^{N} \rho ( f_{\text{photo}} (I_{1}(i, j), I_{1}^{\prime}(i,j)) ).
\label{eq:reconstruction_loss}
\end{equation}
$N$ is the number of pixels. The reconstructed previous frame is computed from the true next frame using inverse warping, $I_{1}^{\prime}(i,j) = I_{2}(i+U_{i,j}, j+V_{i,j})$, accomplished through spatial transformer modules \cite{stn_nips15} inside the CNN. We use a robust convex error function, the generalized Charbonnier penalty $\rho(x) = (x^{2} + \epsilon^{2})^{\alpha}$, to reduce the influence of outliers. $\alpha$ is set to $0.45$. Since unsupervised approaches for optical flow estimation usually fail in regions that are very dark or very bright \cite{jasonUnsup2016}, we choose the tenary census transform \cite{census_transform_Stein_04} as our $f_{\text{photo}}$ to compute the difference between our warped image and the original frame. We adopt it instead of naive photometric differencing as it can compensate for additive and multiplicative illumination changes as well as for changes to gamma, thus providing a more reliable constancy assumption for real imagery. 

This finalizes the design of our forward flow estimation network from the first image to the second. The backward flow from second image to first is estimated using the same model. We now illustrate our occlusion reasoning which uses the forward and backward flow estimates.

\subsection{Occlusion Reasoning}
\label{sec:occlusion}

Occlusion estimation and optical flow estimation are so-called chicken-and-egg problems. In particular, our unsupervised learning framework should not employ the brightness constancy assumption to compute the loss when there is occlusion. Pixels that become occluded in the second frame should not contribute to the photometric error between the true and reconstructed first frames in Eq. \ref{eq:reconstruction_loss}.

We therefore mask occluded pixels when computing the image reconstruction loss in order to avoid learning incorrect deformations to fill the occluded locations. Our occlusion detection is based on the forward-backward consistency assumption \cite{flow_fb_Sundaram_eccv10}. That is, for non-occluded pixels, the forward flow should be the inverse of the backward flow at the corresponding pixel in the second frame. We mark pixels as being occluded whenever the mismatch between these two flows is too large. Thus, for occlusion in the forward direction, we define the occlusion flag $o^{f}$ be 1 whenever the constraint
\begin{equation}
|M^{f} + M^{b}_{M^{f}}|^{2} < \alpha_{1} \cdot (|M^{f}|^{2} + |M^{b}_{M^{f}}|^{2}) + \alpha_{2}
\label{eq:occlusion}
\end{equation}
is violated, and 0 otherwise. $o^{b}$ is defined in the same way, and $M^{f}$ and $M^{b}$ represent forward and backward flow. We set $\alpha_{1}$=0.01, $\alpha_{2}$=0.5 in all our experiments. The resulting occlusion-aware image reconstruction loss is represented as:
\begin{equation}
L = (1 - o^{f}) \cdot L_{\text{reconst}}^{f} + (1 - o^{b}) \cdot L_{\text{reconst}}^{b}
\label{eq:data_loss}
\end{equation}

\begin{figure}[t]
	\centering
	\includegraphics[width=1.0\linewidth,trim=0 0 0 0,clip]{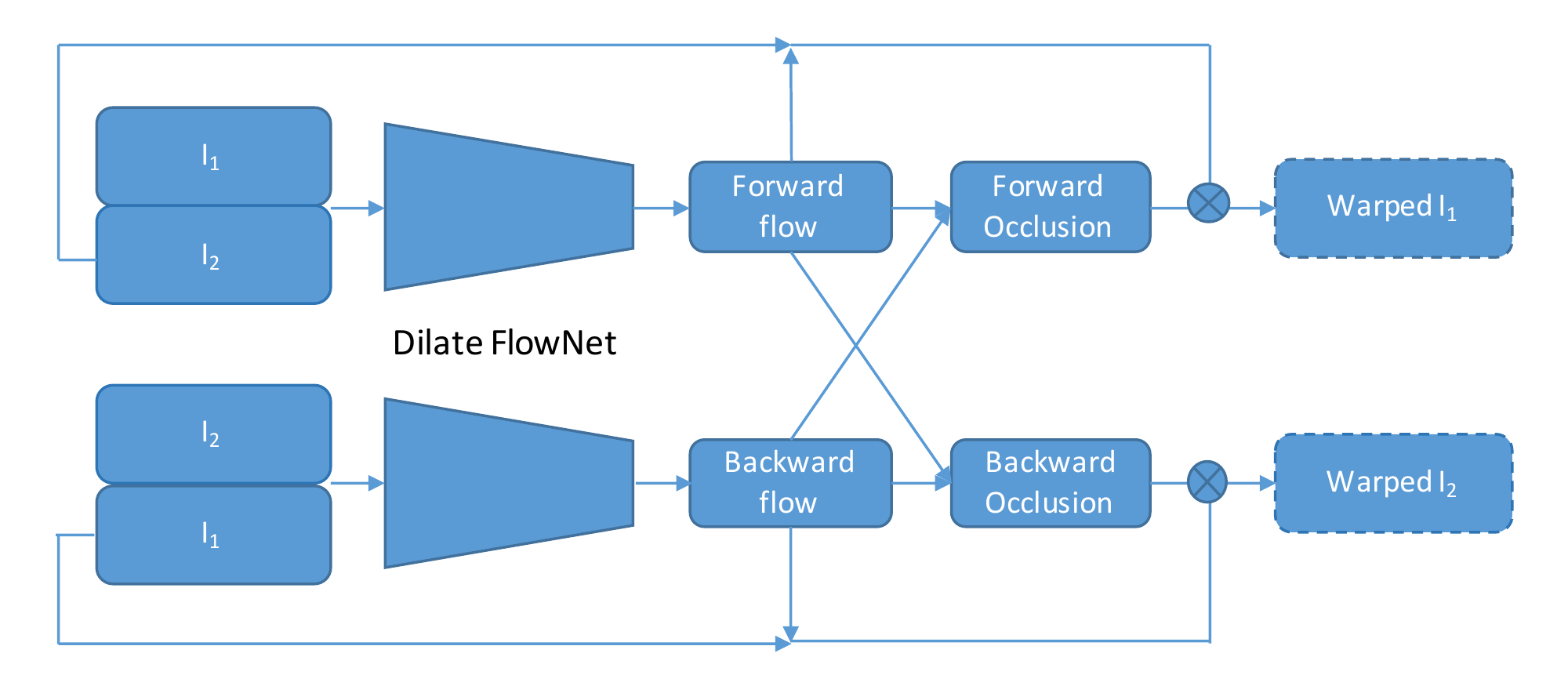}
	\vspace{-4ex}
	\caption{Our bidirectional framework in which the weights are shared between the forward and backward flow estimation. }
	\label{fig:occlusion_overview}
\end{figure}

We also incorporate a second-order smoothness constraint to regularize the local discontinuity of the flow estimation, and a forward-backward consistency penalty on the flow of non-occluded pixels following \cite{unflow_aaai2018}. Our final loss is a weighted sum of all the loss terms. Our full bidirectional framework can be seen in Fig. \ref{fig:occlusion_overview}. 

\section{Experiments}
\label{sec:experiment}

\subsection{Datasets}
\label{sec:datasets}
We train our network in an unsupervised fashion using the SYNTHIA and KITTI raw datasets, and perform our evaluations using the KITTI 2012 and KITTI 2015 benchmarks. Both datasets consist of driving scenes--SYNTHIA is synthetic whereas KITTI is real. For SYNTHIA, we take the left images of the front, back, left, and right views of spring, summer, fall, and winter scenarios from all five sequences, which amounts to 74K image pairs. The KITTI raw dataset contains around 72K image pairs following the split of \cite{jasonUnsup2016,unflow_aaai2018}. 

We first pre-train our network on SYNTHIA without occlusion reasoning since the forward-backward consistency assumption requires reasonably accurate flow estimation. We then fine tune on KITTI with occlusion reasoning. Note that the fine tuning is also unsupervised. During training, we adopt extensive image augmentation, including random scaling, flipping, rotation, Gaussian noise, brightness changes, contrast and gamma changes, and multiplicative color changes. The implementation details can be found in the supplemental materials.

\begin{table}[]
	\centering
	\begin{tabular}{c|cc|cc|c}
		\hline
		KITTI Dataset           & \multicolumn{4}{c|}{2012}     & 2015   \\
		& \multicolumn{2}{l}{AEE(All)} & \multicolumn{2}{c|}{AEE(NOC)} & F1-all \\
		& train        & test       & train       & test     & train                \\
        \hline
        UnsupFlowNet 	& 11.3        & 9.9       & 4.3       & 4.6     & -         \\
        DSTFlow		& 10.43    & 12.4     & 3.29     & 4.0   & 36.0    \\
        DenseNetFlow 	& 10.8        & 11.6       & 3.6       & 4.1  & 36.3 \\
        OcclusionAware 	& $\mathbf{3.55}$        & $\mathbf{4.2}$       & -       & -     & -   \\
        UnFlow-C		& 3.78        & -       & 1.58       & -     & 28.94     \\
        \hline
        Ours			& 3.62        & 4.6       & $\mathbf{1.56}$       & $\mathbf{2.35}$     & $\mathbf{28.45} $   \\
	\end{tabular}
	\vspace{1ex}
	\caption{Comparison of state-of-the-art unsupervised flow estimation on the KITTI 2012 and 2015 benchmarks. AEE(All) and AEE(NOC) are pixel error and F1-all is $\%$ error.}
	\label{tab:flow_results}
    \vspace{-2ex}
\end{table}

\subsection{Results}
\label{sec:results}
We compare our proposed method to recent state-of-the-art unsupervised approaches on the KITTI2012 and KITTI2015 benchmarks. UnsupFlowNet \cite{jasonUnsup2016}, DSTFlow \cite{dstFlow_ren_aaai17}, and DenseNetFlow \cite{densenet_zhu_icip2017} represent earlier approaches to unsupervised optical flow estimation, and are seen to have similar performance. Methods that take occlusion into consideration, OcclusionAware \cite{occlusionFlow_wang17}, UnFlow-C \cite{unflow_aaai2018} and ours, show significant performance improvement. The error rate drops by more $50\%$ on all pixel and non-occluded pixel scenarios. This demonstrates the importance of occlusion reasoning for estimating optical flow especially when the brightness constancy assumption guides the learning.

Note that OcclusionAware \cite{occlusionFlow_wang17} uses the ground truth flow to fine tune the model and so it is not fully unsupervised. UnFlow \cite{unflow_aaai2018} does have other network variants like UnFlow-CSS which can achieve better performance than our method but at a much higher computational cost (three stacked networks). We expect that our model could also benefit from stacking the network multiple times and fine tuning on the ground truth. We show flow visualizations of sample image pairs in the supplemental materials.

\subsection{Generalization}
\label{sec:generalization}

Recent literature \cite{xue17toflow} suggests that endpoint error may not be the best measure of optical flow accuracy especially when the flow is used for other vision tasks. We therefore use the real-world application of action recognition to assess our approach, particularly its ability to generalize to other datasets.

Current state-of-the-art approaches to action recognition are based on two-stream networks \cite{twostream2014,depth2action,dovf_lan_2017}. One stream, termed the spatial stream, operates on the RGB frames. The other stream, termed the temporal stream, operates on the estimated optical flow. Accurate optical flow estimation is key to obtaining good action recognition performance due to its ability to encode human motion information.

We choose UCF101 \cite{ucf101} as our evaluation dataset. We first fine tune our network in an unsupervised fashion on UCF101 to better handle sub-pixel motion. We then train the temporal stream of a standard two-stream action recognition network \cite{twostream2014} using different optical flow estimates, including ours. We use a stack of $5$ optical flow fields as input for fair comparison to \cite{flownet2}. Table \ref{tab:generalization} shows the performance of the action recognition network using the different flow estimates. Ours is shown to result in the highest accuracy and is the fastest. One of the reasons our flow estimates work so well is that our network is unsupervised and so can be fine tuned on tasks for which ground truth flow is not available. In contrast, even though FlowNet2 is a carefully designed system with 5 stacked networks, it is supervised and results in $3\%$ reduced performance. Further, our model can perform inference at real-time rates (i.e., 25fps) which makes it a good candidate for time-sensitive applications. 

\begin{table}[]
	\centering
	\begin{tabular}{c|c|c}
		\hline
		 & Accuracy	(\%) & fps (second)   \\
        \hline
        FlowFields \cite{flow_fields_iccv15}	   & 79.5    & $0.06$ \\
        FlowNet	\cite{flownet}	  & 55.27    & 16.7\\
        FlowNet2	\cite{flownet2}  & 79.51   & 8\\
        \hline
        Ours		  & $\mathbf{82.5}$    & 	$\mathbf{33.3}$	\\
	\end{tabular}
	\vspace{1ex}
	\caption{Two-stream action recognition on the first split of UCF101 using different flow estimates. fps denotes frame per second. }
	\label{tab:generalization}
    \vspace{-2ex}
\end{table}

\section{Conclusion}
\label{sec:conclusion}

We introduce dilated convolution and occlusion reasoning into unsupervised optical flow estimation. The dilated convolution along with dense connectivity preserves the spatial resolution of our estimates without the need to perform upsampling through deconvolution which can result in gridding artifacts. This also reduces the memory footprint of our model, greatly speeding up inference. The occlusion reasoning prevents our network from learning incorrect deformations due to occluded locations. Our approach outperforms state-of-the-art unsupervised approaches to flow estimation on the KTTI benchmark. We also demonstrate its generalization ability to other tasks such as action recognition.

\section{Acknowledgements}
We gratefully acknowledge the support of NVIDIA Corporation through the donation of the Titan X GPUs used in this work.

\bibliographystyle{IEEEbib}
\bibliography{refs}

\begin{thebibliography}{10}

\bibitem{Horn_Schunck}
Berthold~K.P. Horn and Brian~G. Schunck,
\newblock ``{Determining Optical Flow},''
\newblock {\em Artificial Intelligence}, 1981.

\bibitem{opticalFlowWarp2004}
Thomas Brox, Andres Bruhn, Nils Papenberg, and Joachim Weickert,
\newblock ``{High Accuracy Optical Flow Estimation Based on a Theory for
  Warping},''
\newblock in {\em ECCV}, 2004.

\bibitem{brox_flow_matching_11}
T~Brox and J~Malik,
\newblock ``{Large Displacement Optical Flow: Descriptor Matching in
  Variational Motion Estimation},''
\newblock {\em PAMI}, 2011.

\bibitem{flownet}
A.~Dosovitskiy, P.~Fischer, E.~Ilg, P.~Hausser, C.~Hazırbas, V.~Golkov,
  P.~v.d. Smagt, D.~Cremers, and T.~Brox,
\newblock ``{FlowNet: Learning Optical Flow with Convolutional Networks},''
\newblock in {\em ICCV}, 2015.

\bibitem{flownet2}
Eddy Ilg, Nikolaus Mayer, Tonmoy Saikia, Margret Keuper, Alexey Dosovitskiy,
  and Thomas Brox,
\newblock ``{FlowNet 2.0: Evolution of Optical Flow Estimation with Deep
  Networks},''
\newblock in {\em CVPR}, 2017.

\bibitem{spynet_16}
Anurag Ranjan and Michael~J. Black.,
\newblock ``{Optical Flow Estimation using a Spatial Pyramid Network},''
\newblock in {\em CVPR}, 2017.

\bibitem{AhmadiICIP2016}
Aria Ahmadi and Ioannis Patras,
\newblock ``{Unsupervised Convolutional Neural Networks for Motion
  Estimation},''
\newblock in {\em ICIP}, 2016.

\bibitem{jasonUnsup2016}
Jason~J. Yu, Adam~W. Harley, and Konstantinos~G. Derpanis,
\newblock ``{Back to Basics: Unsupervised Learning of Optical Flow via
  Brightness Constancy and Motion Smoothness},''
\newblock in {\em ECCV Workshop}, 2016.

\bibitem{hidden_zhu_17}
Yi~Zhu, Zhenzhong Lan, Shawn Newsam, and Alexander~G. Hauptmann,
\newblock ``{Hidden Two-Stream Convolutional Networks for Action
  Recognition},''
\newblock {\em arXiv preprint arXiv:1704.00389}, 2017.

\bibitem{guided_zhu_2017}
Yi~Zhu, Zhenzhong Lan, Shawn Newsam, and Alexander~G. Hauptmann,
\newblock ``{Guided Optical Flow Learning},''
\newblock in {\em CVPR Workshop}, 2017.

\bibitem{Lai-NIPS-2017}
Wei-Sheng Lai, Jia-Bin Huang, and Ming-Hsuan Yang,
\newblock ``{Semi-Supervised Learning for Optical Flow with Generative
  Adversarial Networks},''
\newblock in {\em NIPS}, 2017.

\bibitem{occlusionFlow_wang17}
Yang Wang, Yi~Yang, Zhenheng Yang, Liang Zhao, and Wei Xu,
\newblock ``{Occlusion Aware Unsupervised Learning of Optical Flow},''
\newblock {\em arXiv preprint arXiv:1711.05890}, 2017.

\bibitem{unflow_aaai2018}
Simon Meister, Junhwa Hur, and Stefan Roth,
\newblock ``{UnFlow: Unsupervised Learning of Optical Flow with a Bidirectional
  Census Loss},''
\newblock in {\em AAAI}, 2018.

\bibitem{dilate_Yu_iclr16}
Fisher Yu and Vladlen Koltun,
\newblock ``{Multi-Scale Context Aggregation by Dilated Convolutions},''
\newblock in {\em ICLR}, 2016.

\bibitem{densenet_16}
Gao Huang, Zhuang Liu, Kilian~Q. Weinberger, and Laurens van~der Maaten,
\newblock ``{Densely Connected Convolutional Networks},''
\newblock in {\em CVPR}, 2017.

\bibitem{densenet_zhu_icip2017}
Yi~Zhu and Shawn Newsam,
\newblock ``{DenseNet for Dense Flow},''
\newblock in {\em ICIP}, 2017.

\bibitem{guided_filter_He_eccv10}
Kaiming He, Jian Sun, and Xiaoou Tang,
\newblock ``{Guided Image Filtering},''
\newblock in {\em ECCV}, 2010.

\bibitem{odena2016deconvolution}
Augustus Odena, Vincent Dumoulin, and Chris Olah,
\newblock ``Deconvolution and checkerboard artifacts,''
\newblock {\em Distill}, 2016.

\bibitem{drn_Yu_cvpr17}
Fisher Yu, Vladlen Koltun, and Thomas Funkhouser,
\newblock ``{Dilated Residual Networks},''
\newblock in {\em CVPR}, 2017.

\bibitem{stn_nips15}
Max Jaderberg, Karen Simonyan, Andrew Zisserman, and Koray Kavukcuoglu,
\newblock ``{Spatial Transformer Network},''
\newblock in {\em NIPS}, 2015.

\bibitem{census_transform_Stein_04}
Fridtjof Stein,
\newblock ``{Efficient Computation of Optical Flow Using the Census
  Transform},''
\newblock in {\em DAGM Pattern Recognition}, 2004.

\bibitem{flow_fb_Sundaram_eccv10}
Narayanan Sundaram, Thomas Brox, and Kurt Keutzer,
\newblock ``Dense point trajectories by gpu-accelerated large displacement
  optical flow,''
\newblock in {\em ECCV}, 2010.

\bibitem{dstFlow_ren_aaai17}
Zhe Ren, Junchi Yan, Bingbing Ni, Bin Liu, Xiaokang Yang, and Hongyuan Zha,
\newblock ``{Unsupervised Deep Learning for Optical Flow Estimation},''
\newblock in {\em AAAI}, 2017.

\bibitem{xue17toflow}
Tianfan Xue, Baian Chen, Jiajun Wu, Donglai Wei, and William~T Freeman,
\newblock ``Video enhancement with task-oriented flow,''
\newblock {\em arXiv preprint arXiv:1711.09078}, 2017.

\bibitem{twostream2014}
K.~Simonyan and A.~Zisserman,
\newblock ``{Two-Stream Convolutional Networks for Action Recognition in
  Videos},''
\newblock {\em NIPS}, 2014.

\bibitem{depth2action}
Yi~Zhu and Shawn Newsam,
\newblock ``{Depth2Action: Exploring Embedded Depth for Large-Scale Action
  Recognition},''
\newblock in {\em ECCV Workshops}, 2016.

\bibitem{dovf_lan_2017}
Zhenzhong Lan, Yi~Zhu, Alexander~G. Hauptmann, and Shawn Newsam,
\newblock ``{Deep Local Video Feature for Action Recognition},''
\newblock in {\em CVPR Workshops}, 2017.

\bibitem{ucf101}
Khurram Soomro, Amir~Roshan Zamir, and Mubarak Shah,
\newblock ``{UCF101: A Dataset of 101 Human Action Classes From Videos in The
  Wild},''
\newblock in {\em CRCV-TR-12-01}, 2012.

\bibitem{flow_fields_iccv15}
Christian Bailer, Bertram Taetz, and Didier Stricker,
\newblock ``{Flow Fields: Dense Correspondence Fields for Highly Accurate Large
  Displacement Optical Flow Estimation},''
\newblock in {\em ICCV}, 2015.

\end{thebibliography}

\end{document}